\documentclass[runningheads]{llncs}
\usepackage[T1]{fontenc}
\usepackage{graphicx}
\usepackage{color}
\usepackage{amsmath}
\usepackage{multirow}
\usepackage{booktabs} 
\usepackage{tabularx}
\usepackage{hyperref}
\usepackage{xcolor}

\begin{document}
\title{Quantitative Evaluation of the Saliency Map for Alzheimer's Disease Classifier with Anatomical Segmentation}
\titlerunning{Quantitative Evaluation of the Saliency Map for AD with Segmentation}
%
\author{Yihan Zhang\inst{1} \and Xuanshuo Zhang\inst{1} \and 
Wei Wu\inst{1} \and
Haohan Wang\inst{2}}
\authorrunning{Zhang et al.}
%
\institute{Carnegie Mellon University \\
\email{weiwu2@cs.cmu.edu}\\
 \and
University of Illinois Urbana-Champaign
 \\
\email{haohanw@illinois.edu}}
\maketitle 
\begin{abstract}
Saliency maps have been widely used to interpret deep learning classifiers for Alzheimer's disease (AD). 
However, since AD is heterogeneous and has multiple subtypes, the pathological mechanism of AD remains not fully understood and may vary from patient to patient. Due to the lack of such understanding, it is difficult to comprehensively and effectively assess the saliency map of AD classifier. In this paper, we utilize the anatomical segmentation to allocate saliency values into different brain regions. By plotting the distributions of saliency maps corresponding to AD and NC (Normal Control), we can gain a comprehensive view of the model's decisions process. In order to leverage the fact that the brain volume shrinkage happens in AD patients during disease progression, we define a new evaluation metric, brain volume change score (VCS), by computing the average Pearson correlation of the brain volume changes and the saliency values of a model in different brain regions for each patient. Thus, the VCS metric can help us gain some knowledge of how saliency maps resulting from different models relate to the changes of the volumes across different regions in the whole brain. We trained candidate models on the ADNI dataset and tested on three different datasets. Our results indicate: (i) models with higher VCSs tend to demonstrate saliency maps with more details relevant to the AD pathology, (ii) using gradient-based adversarial training strategies such as FGSM and stochastic masking can improve the VCSs of the models. Our source code is available at \href{https://github.com/yihanz3/EvaADSaliency}{GitHub}. 

\end{abstract}
\section{Introduction}
Interpretability is a crucial factor driving artificial intelligence models towards clinical applications~\cite{zhang2021survey,liu2023towards}. 
Saliency map is among the most commonly used tools for interpreting medical imaging models and has been widely applied in Alzheimer's Disease prediction tasks~\cite{Zhang2022Overlook,gradcam2023}. It helps us understand the model's decision-making process by visualizing the contribution of each part of the image to the model's prediction. 
A model is considered effective when, in the presence of defined target regions, areas of high saliency significantly overlap with these target regions~\cite{Zhou2021GANlossAD}.
For instance, in an effective ``plaque'' recognizer, high saliency regions should have high overlap rate with the pixels occupied by the ``plaque''. However, due to the lack of a unified conclusion on the pathogenesis of AD and the proven existence of multiple subtypes, it is challenging to perform an effective and comprehensive evaluation of AD saliency maps~\cite{Yin2023PNAS}.
\begin{figure}[h]
    \centering
    \includegraphics[width=1\textwidth]{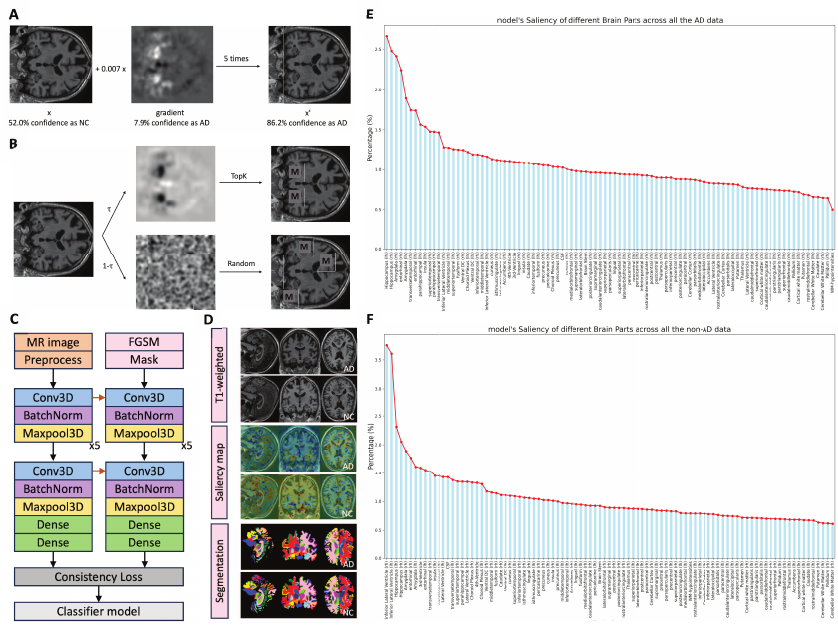}
    \caption{\textbf{A}. Fast Gradient Sign Method(FGSM). After adding perturbation several times, NC patient is then misclassified by the target network as AD when it is still clearly NC. \textbf{B}.Stochastic Masking. For each image, model has a probability of $\tau$ to be masked by TopK blocks with highest gradients, while 1-$\tau$ to be masked with random blocks. \textbf{C}.Architecture of the training framework. \textbf{D}. T1-weighted MR image, Saliency map, and Segmentation map in x-y-z axis. \textbf{E}. The averaged distribution of model's gradient across brain regions for all AD patients, computed by overlapping of Saliency map and segmentation map. \textbf{F}. The averaged distribution of model's gradient across brain regions for all non-AD patients.}
    \
    \label{fig:model}
\end{figure}

For a long time, interpreting AD’s saliency maps has primarily focused on atrophy to the hippocampus~\cite{Zhou2021GANlossAD}, as it is crucial for memory formation and is one of the initial areas of the brain to be affected. However, increasing evidence suggests that focusing solely on the hippocampus is insufficient, as the impact of AD actually extends across the entire brain~\cite{fri2010,rosen2010}. Global brain atrophy, including extensive damage to subcortical structures and the cortex, is a hallmark of AD progression~\cite{oss2016,firtg2019}. This comprehensive damage explains why patients with AD experience a full spectrum of decline, from simple memory loss to severe cognitive impairments. Therefore, the ideal evaluation method for saliency maps should utilize information from the entire brain.

In this paper, we propose a metric for evaluating saliency maps that aims to analyze information from the entire brain, named the brain volume change score (VCS).
By utilizing anatomical segmentation, we can allocate saliency values to different brain regions. 
This approach acknowledges the extensive and variable impact of AD on different brain regions, underpinning the necessity for a comprehensive evaluation that mirrors the disease's global brain atrophy, thereby enabling a more nuanced understanding of how well the saliency maps reflect the actual brain volume changes observed in AD progression.
We then use Pearson correlation to bridge the model’s saliency values and the actual volume change for each brain region. VCS is computed by averaging the Pearson correlation for all patients. 
Volume changes are constant for specific dataset, thus VCS is then the metric only dependent on model's saliency value, which is computed from model's gradient.

In addition, we seek further for methods that improve VCS. Gradient-based adversarial training strategies are widely used in machine learning community as data augmentation and model's robustness~\cite{Geceri2019grad-based,Madry2017Adversarial}, and proven to improve the interpretability of saliency map~\cite{andr2020inter-adv}. In this paper, we apply two of these methods, Fast Gradient Sign Method (FGSM)~\cite{Geceri2019grad-based} and gradient-based stochastic masking~\cite{chen2023masked,huang2022spectrum,xie2022simmim}, to train the model, as shown in Fig.~\ref{fig:model}. Our experiments demonstrate that these two gradient-based training strategies can improve the model's VCS compared to the baseline 3D convolutional network.

\section{Method}
\subsubsection{Plot the distribution of the model’s gradient across brain region}
We use the FastSurfer~\cite{fastsurfer2020} to segment the original image into 95 brain regions automatically. Using these segmentation maps $C_{seg, i}$ = $\{c_{n, i}\} ^{N}_{n=1}$, we then compute the overlapped saliency value $S_{n,i}$ by collecting the gradients $\nabla_x J(\theta, x_{i}, y_{i})$ within each brain region $n$ for each patient $i$. Then we normalize each saliency value with corresponding size of the segmentation mask $V_{n,i}$. By doing this, we can plot the distribution of model's gradients across all the brain regions, as shown in Fig.~\ref{fig:model}.
\subsubsection{Brain volume change score}
In order to better investigate the similarity of model's decision with actual happening to the patients, we define brain volume change score (VCS) as a quantitative evaluation of saliency map's Interpretability in Alzheimer's disease, by calculating Pearson correlation $\mathcal{P}_{i}$ between the overlapped saliency value $S_{n,i}$ and the actual shrinkage volume for all different brain regions $\Delta V_{n,i}$ for each patient. Then, VCS is computed as the average of $\mathcal{P}_{i}$:
\[
VCS = \frac{1}{\text{I}}\sum_{i=1} \mathcal{P}_{i} =  \frac{1}{\text{I}}\sum_{i=1} \frac{\sum_{n=1}^{N} (S_{n,i} - \bar{S_{i}})(\Delta V_{n,i} - \bar{\Delta V_{i}})}{\sqrt{\sum_{n=1}^{N} (S_{n,i} - \bar{S_{i}})^2}\sqrt{\sum_{n=1}^{N} (\Delta V_{n,i}- \bar{\Delta V_{i}})^2}}
\]
Here, $i$ denotes  $i-th$ patient, I denotes total number of patients, $\Delta V_{n,i}$ is computed from volume change between $V_{n,i}^{t_{i}=T_{i}}$ and $V_{n,i}^{t_{i}=0}$, $t_{i} \in \{0, 1, \dots, T_{i}\}$ denotes the time when patient $i$ visited the hospital. $N$ denotes the total number of brain regions, in our case $N$=95.

\subsubsection{Fast Gradient Sign Method}
Given a neural network represented by a function \( f \) with parameters \( \theta \), mapping an input \( x \) to an output, and a cross-entropy loss function \( J(\theta, x, 1-y) \) where \( y \in {0,1}\) is the true label, \( 1-y \) denotes the adverse label. The goal of FGSM is to create an adversarial example \( x_{adv} \) that maximizes \( J \) subject to \( \|x_{adv} - x\|_p \leq \epsilon \). FGSM algorithm can be writen as:
\[ x'_{t+1} = x_t + \alpha \cdot \text{sign}(\nabla_x J(\theta, x_t, 1-y)) \]
Update the adversarial example by moving in the direction of the sign of the gradient, scaled by a step size \( \alpha \) for \( t = 0, \ldots, T-1 \), then clip the  adversarial example \( x'_{t+1} \) within the $\epsilon$-constrained space.

\subsubsection{Consistency loss}
We apply a consistency loss function \( L_{con} \) ~\cite{Balar2022Consistency}that operates on the gradients of the model. Given a pair of inputs original image \( x \) and post-FGSM image \( x' \), the consistency loss is calculated as the difference in the gradients of our model's output with respect to these inputs. Formally, this can be expressed as:
\[ L_{con} = J(\theta, x, y) + J(\theta, x', y) + \lVert \nabla_x J(\theta, x, y) - \nabla_{x'} J(\theta, x', y) \rVert^2_2 \]

where \( J(\theta, x, y) \) is the loss function of the model with parameters \( \theta \), \( \nabla_x \) and \( \nabla_{x'} \) denote the gradients with respect to \( x \) and \( x' \) respectively, and \( \lVert \cdot \rVert_2 \) represents the $L_{2}$-norm, \( \lambda \) is a regularization parameter that balances the primary loss and the consistency loss. By minimizing \( L_{con} \), the model is encouraged not only to perform well on the primary task but also to maintain consistent gradients, thereby enhancing its robustness to input perturbations and ensuring reliable interpretability.

\subsubsection{Stochastic Masking}
We introduce a stochastic masking strategy, as shown in Fig.~\ref{fig:model}, applied during the later stage of training process to achieve two primary objectives: enhancing the gradient robustness of our model and encouraging the model to reason across the whole brain, thereby mitigating the risk of overfitting to a few decisive factors. Initially, each image \( x \) is partitioned into 6×6×6 small blocks. During each epoch of the later training process, we mask the image with \( p\% \) of these blocks. The proportion \( p \) is uniformly sampled between 10 to 40. The selection of blocks to be masked also follows a stochastic approach governed by a parameter \( \tau \). For each image \( x \), with probability \( \tau \), blocks are selected randomly, while with probability \( 1 - \tau \), we employ a gradient-based greedy masking approach. For this greedy masking, we compute overlapped gradient for each block and select the top-\( k \)~\cite{jayakumar2020topkast} blocks for the masking.

\section{Experiments and Results}
    \begin{figure}[h]
    \centering
    \includegraphics[width=1\textwidth]{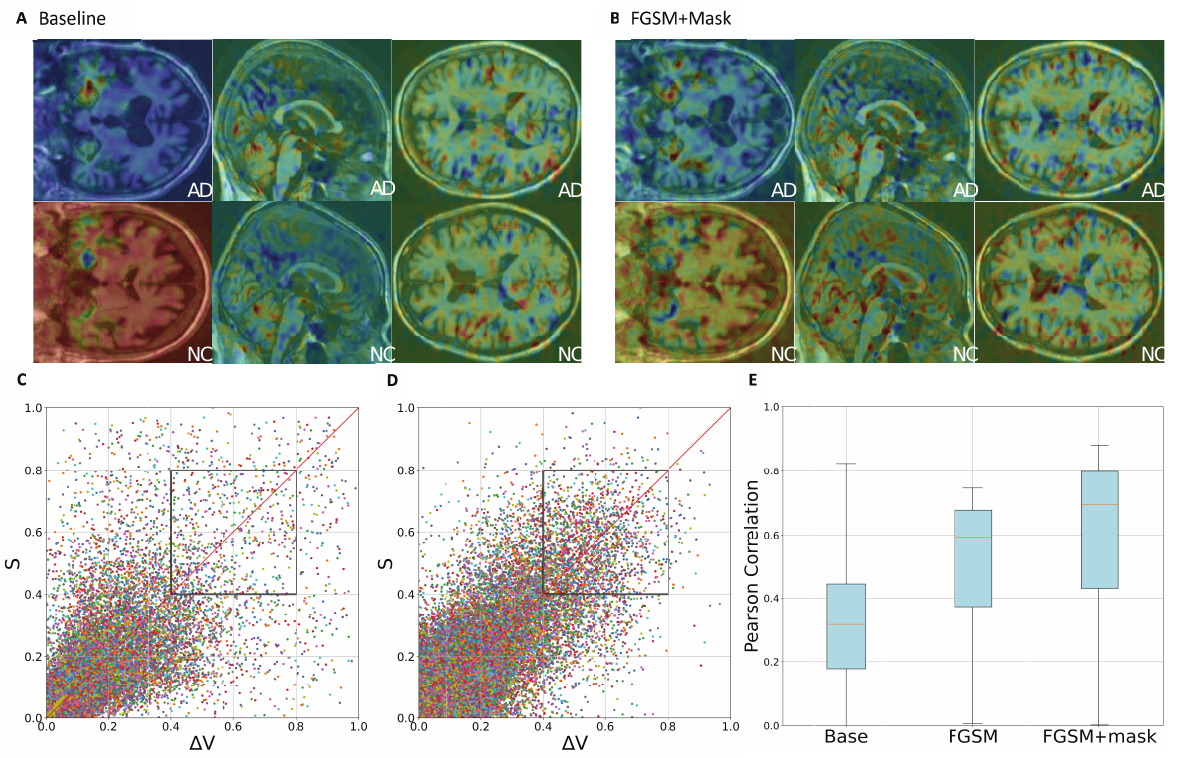}
    \caption{\textbf{A}. Saliency map of Baseline 3d-CNN. \textbf{B}. Saliency map of 3d-CNN+FGSM+mask. \textbf{C}. Scatter plot of $\Delta V_{n,i}$ vs $S_{n, i}$ of baseline model. \textbf{D}. Scatter plot of $\Delta V_{n,i}$ vs $S_{n, i}$ of FGSM+mask model. We can see a higher correlation (red line denotes correlation equals 1) especially when $\Delta V_{n,i}$ is high (shown in black box). \textbf{E}. Box plot of all $\mathcal{P}_{i}$ for three candidate models. }
    \
    \label{fig:eva-smap}
\end{figure}

\subsubsection{Data Preprocessing}
Our data was collected from four different data collections for training and cross-dataset tests. We train our model on ADNI subset, which consists of follow-up data with several sessions, as 336 subjects with 1106 samples with AD category and 330 subjects with 1830 samples with NC category. We test our model on MIRIAD (23 NC, 46 AD); OASIS (605 NC, 493 AD); AIBL (429 NC, 76 AD). 

We use the t1-linear pipeline of Clinica~\cite{clinicadl2022} to register images, which were further cropped to remove the background resulting in images of size $169\times208\times179$, with 1 mm isotropic voxels. Overall, the registration process we perform on the data maps different sets of images into a single coordinate system.

\subsubsection{Saliency map Interpretation}
We plot the distribution of model's saliency value (gradient) across the different brain regions to see which regions contribute most to model's decision. As shown in Fig.~\ref{fig:model} E and F, the result shows that hippocampus contributes most for model to decide this patient gets AD, while inferior lateral ventricle contributes most for NC. Comparing the difference between distribution of AD and NC, we found that regions contributing most to the model's differentiation include: hippocampus, the inferior lateral ventricle, the lateral ventricle, the amygdala, the 3rd ventricle, the hippocampus, and the cerebrospinal fluid (CSF). To assess the saliency map comprehensively, We then compute the VCS for three candidate models, as shown in Fig.~\ref{fig:eva-smap}. Our results show that FGSM+mask has more detailed saliency map than baseline 3d-CNN, thus having higher VCS. Through scatter plot of $\Delta V_{n,i}$ vs $S_{n, i}$, we found the ability of learning information associating with higher $\Delta V_{n,i}$ might be the reason to explain why FGSM+mask can perform more trustworthy saliency map.

\subsubsection{Classification Results}
We train and validate our model based on the ADNI dataset as a two-class classification task (NC vs AD). 
Four evaluation metrics are used, including brain volume change score (VCS), accuracy (ACC), sensitivity (SEN), and specificity (SPE). We compare 3d-CNN+FGSM+mask (Mask for short) with the traditional 3d-CNN, and 3d-CNN+FGSM (FGSM for short). For all the comparison settings, we set the same training parameters. As shown in the Table.~\ref{tab}, although there might be a slight decrease in SPE, applying FGSM and stochastic masks can both improve the model's VCS, accuracy, and SEN compared with baseline 3d-CNN, especially for OASIS3 dataset. Since AIBL, MIRIAD and OASIS3 don't have follow-up data on disease progression as ADNI do, we only compute VCS for ADNI with three candidate models.

\begin{table}[htbp]
\centering
\caption{Cross-dataset classification results}
\begin{tabularx}{\textwidth}{lccccccccccccccc}
\toprule
 & \multicolumn{3}{c}{VCS} 
 & \multicolumn{3}{c}{ACC} 
 & \multicolumn{3}{c}{SPE} 
 & \multicolumn{3}{c}{SEN} \\
\cmidrule(r){2-4} \cmidrule(r){5-7} \cmidrule(r){8-10} 
\cmidrule(r){11-13}
 & Base & FGSM & Mask 
 & Base & FGSM & Mask 
 & Base & FGSM & Mask 
 & Base & FGSM & Mask 
 \\
\midrule
ADNI
& 0.32  & 0.51  & \textbf{0.60} 
& 85.70 & 88.24 & \textbf{90.20} 
& 93.76 & \textbf{94.21} & 92.25 
& 61.75 & 73.58 & \textbf{88.44} 
\\
AIBL  
& - & - & - 
& 88.62 & 88.79 & \textbf{90.31} 
& \textbf{97.24} & 94.33 & 92.86 
& 27.45 & 74.66 & \textbf{81.43} 
\\
MIRIAD   
& - & - & - 
& 81.84 & 86.01 & \textbf{90.56} 
& \textbf{98.30} & 96.42 & 93.13 
& 73.49 & 83.72 & \textbf{87.57} 
\\
OASIS3 
& - & - & - 
& 80.44 & 81.25 & \textbf{88.15} 
& \textbf{95.98} & 91.55 & 89.14 
& 23.71 & 43.32 & \textbf{87.64} 
\\
\bottomrule
\end{tabularx}
\label{tab}
\end{table}

\section{Conclusions}
    We proposed Volume Change Score to evaluate the saliency map of Alzheimer’s Disease with anatomically segmentation. The VCS demonstrates the information of model's decision with the basic facts of volume change of brain regions. It shows potential as a criterion to evaluate the saliency map, although only available to be computed when dataset (e.g. ADNI) contains follow-up data. Besides, we apply a series of gradient-based training strategies, which shows success to improve the VCS of AD classifiers and is easy to implement in other medical problems.

\section{Notes}
*Data used in preparation of this article were obtained publicly from the Alzheimer’s Disease Neuroimaging Initiative (ADNI) database (adni.loni.usc.edu), Minimal Interval Resonance Imaging in Alzheimer's Disease (MIRIAD) database (nitrc.org/projects/miriad), Open Access Series of Imaging Studies (OASIS) database (oasis-brains.org) and Australian Imaging, Biomarker Lifestyle Flagship Study of Ageing (AIBL) database (sunbirdbio.com).

%
%
%
%

\end{document}